# Framework for Robust Motion Planning of Tethered Multi-Robot Systems in Marine Environments

Markus Buchholz[1], Ignacio Carlucho[1], Zebin Huang[2], Michele Grimaldi[1], Pierre Nicolay[2], Sümer Tunçay[1] and Yvan R. Petillot[1]

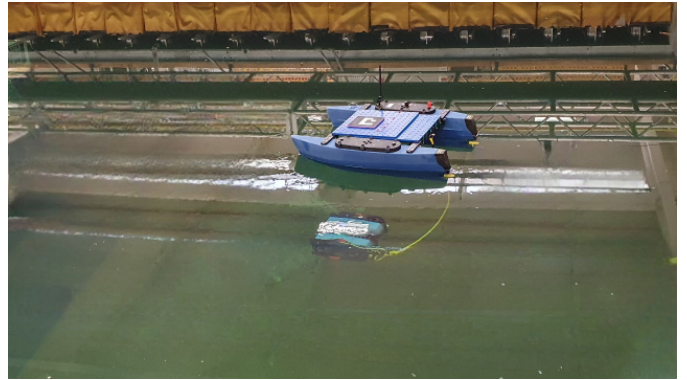

Fig. 1: Experimental setup showcasing the deployment of ASV and AUV constrained by catenary tether in yellow, in a controlled flooded environment for testing.

*Abstract*— This paper introduces CoralGuide, a novel framework designed for path planning and trajectory optimization for tethered multi-robot systems. We focus on marine robotics, which commonly have tethered configurations of an Autonomous Surface Vehicle (ASV) and an Autonomous Underwater Vehicle (AUV). CoralGuide provides safe navigation in marine environments by enhancing the A* algorithm with specialized heuristics tailored for tethered ASV-AUV systems. Our method integrates catenary curve modelling for tether management and employs Bezier curve interpolation for smoother trajectory planning, ensuring efficient and synchronized operations without compromising safety. Through simulations and real-world experiments, we have validated CoralGuide's effectiveness in improving path planning and trajectory optimization, demonstrating its potential to significantly enhance operational capabilities in marine research and infrastructure inspection.

## I. INTRODUCTION

In recent years marine robots have become increasingly important for offshore industry, such as inspection of offshore windfarms [1]. To extend mission time and mitigate the communication issues of underwater vehicles, a combination of Autonomous Underwater Vehicles (AUVs) and Autonomous surface vehicles (ASVs) are used [2]. Typically these vehicles are connected by a tether, where the ASV provides the AUV with power and a communication link via the umbilical. However, in this type of setups, tether management plays a critical role, as unmanaged tethers can lead to entanglements [3], [4]. The consequences of this can range from substantial delays in the operation, requiring human operators to untangle the tether, to line cuts with the potential of loss or damage to the Remotely Operated Vehicles (ROVs). Therefore it is necessary to ensure coordinated operation and obstacle avoidance, for both vehicles, while ensuring safe tether management.

Conventional methods for navigating often struggle with the unique operational demands and the dynamic nature of tethered ASV-AUV systems [5], [6], [7], [8]. The tether acts as an additional constraint to the multi-robot systems, but that is not controllable. These difficulties are magnified in underwater environments due to the highly non-linear dynamics, environmental disturbances, and limited sensor capabilities

This research was supported by the EPSRC Centre for Doctoral Training in Robotics and Autonomous Systems under the Grant Reference EP/S023208/1.

[1] School of Engineering & Physical Sciences, Heriot-Watt University, Edinburgh, UK m.buchholz@hw.ac.uk
[2] School of Mathematics & Computer Sciences, Heriot-Watt University, Edinburgh, UK

[9]. Furthermore, disturbances in marine environments can also affect the tether, which increases the problem difficulty [10].

To address the challenges of tethered ASV-AUV system navigation, our study introduces *CoralGuide*. The code of our framework is open-sourced[1]. *CoralGuide* refines A* with heuristics tailored to the unique constraints of underwater tethered systems. This advancement facilitates path planning by integrating catenary curve modelling for effective tether management. By systematically accounting for the tether's influence on movement and incorporating these factors into the motion planning process, *CoralGuide* aims to improve operational coherence between the ASV and AUV, ensuring safe and synchronized manoeuvres in a domain where precision is critical. Fig. 1 demonstrates our research system, illustrating how two AUVs, connected by tethers, are deployed in a test environment. We provide results both in simulation and in a real-world environment to showcase CoralGuide's performance. Our results show that our method is able to plan a trajectory for both vehicles while ensuring a collision-free trajectory and disentanglement of the tether, thus allowing for safe mission execution.

The key contributions of this study are:
- An effective path planning and trajectory optimization framework for tethered ASV-AUV systems, which enhances the reliability and practicality of navigating such setups.
- A novel approach to synchronizing the movements of

[1] https://github.com/markusbuchholz/marine-robotics-sim-framework

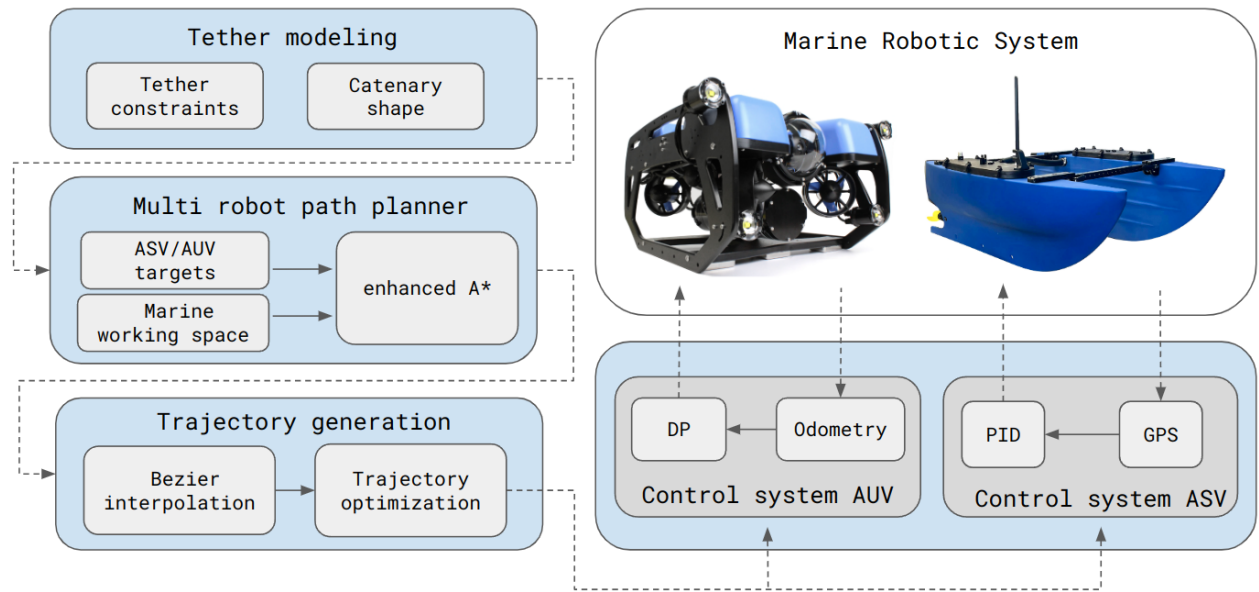

Fig. 2: System overview of a tethered ASV-AUV collaboration framework for autonomous navigation.

ASVs and AUVs, ensuring they reach their waypoints together and maintain the integrity of the tether system.
- The first method to reliably solve the tether multi-robot planning problem in a realistic underwater environment.

## II. RELATED WORK

The problem of multi-robot motion planning has been studied extensively [11], [12]. Particularly, in the underwater domain, some initial works have developed path planners for a fleet of underwater vehicles using the rapidly exploring random trees star (RRT*) algorithm [13]. Other works have focused on the problem of path planning of an ASV working together with a fleet of AUVs to improve the localization accuracy [14]. However, the majority of these works often overlook the complexities introduced by tether systems.

Recent works have started to explore the problem of multi-robot motion planning in tethered systems, but these are mostly focused on aerial vehicles [15]. In [16] the authors present a non-linear optimization method for tethered aerial robots. An extension to this work was presented in [17], which develops a path planner for a tethered ground-aerial robot system. The path planner focuses on achieving a fixed goal for the aerial vehicle, while the ground vehicle does not have a fixed goal and is free to position itself on the map. The approach in [18], utilizes braid theory for multiple tethered robots, and introduces a novel perspective on avoiding entanglements, yet its application is limited to aerial drones. Similarly, [19] focuses on multi-robot tether-aware path planning, but on systems that are tethered to a fixed base station. Overall, while these works present notable advances, they do not take into consideration the complex dynamics of marine systems [20].

Recent studies have begun addressing tether management for ROVs and tethered ASV-AUV systems through both passive and active strategies to mitigate entanglement risks and enhance operational efficacy [21], [22], [23]. A visual servoing method for real-time tether control of a set of linked AUVs is presented in [24]. In [25], the authors explore a catenary shape estimation for AUVs connected in a chain. Similarly, our work makes use of the catenary tether model to prevent entanglements, but it incorporates them as a constraint into our motion planning systems, which allows us to safely plan and execute motions in an ASV-AUV system.

## III. PROBLEM FORMULATION

We consider a system comprising an ASV and AUV, each defined by start and goal positions in a three-dimensional marine environment. The ASV, operating at the water's surface, is represented by $p_{\text{ASV}}(t) = (x_{\text{ASV}}(t), y_{\text{ASV}}(t), z_{\text{ASV}})$, where $z_{\text{ASV}}$ is constant due to its surface operation. The AUV's position is denoted as $p_{\text{AUV}}(t) = (x_{\text{AUV}}(t), y_{\text{AUV}}(t), z_{\text{AUV}}(t))$, reflecting its capability for three-dimensional navigation. These vehicles are connected by a tether of fixed length $l$, necessitating the constraint $l \geq \|p_{\text{AUV}}(t) - p_{\text{ASV}}(t)\|$ at all times to prevent tension that could damage the tether.

The objective of the CoralGuide path planner is to devise an obstacle-free trajectory for the ASV, AUV, and the tether, guiding each from their respective starting points to their designated goals while satisfying the tether's length constraints. Formally, the task is formulated as follows:

$$\min_{p_{\text{ASV}}(t), p_{\text{AUV}}(t)} \quad \text{subject to:}$$

1) An obstacle-free path is maintained for $p_{\text{ASV}}(t)$, $p_{\text{AUV}}(t)$, and the tether over the interval $t \in [0, T]$.
2) The tether length constraint $l \geq \|p_{\text{AUV}}(t) - p_{\text{ASV}}(t)\|$ is upheld for all $t \in [0, T]$,

where $T$ denotes the time required to reach the goal positions. This ensures that both vehicles and their connecting tether navigate safely and efficiently through the marine working space, from start to goal, within the operational limits defined by the tether's length.

## IV. METHOD DESCRIPTION

### A. System Overview

In this section, we provide an overview of our proposed methodology for a motion planning method that allows coordination between tethered ASV and AUV systems. The architecture and main components of CoralGuide are demonstrated in Fig. 2.

The architecture comprises four core modules: Tether Modeling, Multi-Robot Path Planning, Trajectory Generation, and Control. The Tether Modeling module mimics the tether's natural catenary shape in 3D space, considering its length, tensile strength, and flexibility to prevent entanglement and ensure realistic behaviour. The Path Planning module utilizes a refined A* algorithm, focusing on maintaining the tether's catenary shape while navigating obstacle-free paths for the ASV, AUV, and tether. This is achieved by incorporating heuristic strategies that account for the tether's physical constraints, ensuring optimal routing without stress or entanglement risks. Trajectory Generation employs Bezier Interpolation, fine-tuned through optimization to create smooth paths that respect the tether's physical behaviour, facilitating smooth and feasible paths. The Control module utilizes a combination of Proportional-Integral-Derivative (PID) control strategies to achieve precise navigation and stabilization of the vehicles.

### B. Tether Modeling

The tether is modelled as lumped masses at discretized points along a catenary curve, defined by the vector function $\vec{r}(t) = (x(t), y(t), z(t))$ for $t \in [0, 1]$. This curve is parameterized for $N$ equidistant points as follows:

$$\vec{r}_i = f\left(\frac{i}{N-1}\right), \quad i = 0, \ldots, N-1,$$

where $\vec{r}_i$ represents the position of the $i^{th}$ lumped mass along the tether. To describe the 3D catenary curve that connects the ASV and AUV, we extend the classical 2D catenary equation into three dimensions.

Let $d = \sqrt{(x_2 - x_1)^2 + (y_2 - y_1)^2}$ represent the planar distance between the start and end points of the tether. Then, the positions $x(t)$, $y(t)$, and $z(t)$ of a point on the tether's catenary curve in 3D space, parameterized by $t \in [0, 1]$, are given by:

$$\begin{cases} d = \sqrt{(x_2 - x_1)^2 + (y_2 - y_1)^2}, \\ x(t) = x_1 + (x_2 - x_1) \cdot t, \\ y(t) = y_1 + (y_2 - y_1) \cdot t, \\ z(t) = a \cosh\left(\frac{d \cdot (t - 0.5)}{a}\right) + z_0 - a \cosh\left(\frac{d}{2a}\right), \end{cases}$$

where $a$ is the catenary constant, $z_0$ is the vertical offset ensuring the curve passes through the desired midpoint in the vertical direction, and $t$ interpolates between the start and end points of the tether.

Given the initial path determined by CoralGuide's A* algorithm for one vehicle, the path for the second vehicle is computed considering the tether's catenary characteristics to maintain a safe and viable configuration throughout the motion. This ensures the generated paths are optimal and collision-free for both vehicles and also the tether.

### C. Path planning

We formulate the path planning problem for a system consisting of an ASV and an AUV tethered together, incorporating orientation as a critical parameter in addition to their spatial positions. The ASV and AUV navigate through a 3D marine environment, aiming to reach predefined goal positions while avoiding obstacles and managing the physical constraints imposed by the tether.

The objective is to find paths for the ASV and AUV, denoted as $P_{\text{ASV}}(t)$ and $P_{\text{AUV}}(t)$, that navigate through a marine working space from start to goal positions while ensuring the tether maintains its catenary shape and avoiding obstacles. Particularly, in our formulation the problem can be described as:

Find $P_{\text{ASV}}(t), P_{\text{AUV}}(t)$,

s.t. $\begin{cases} \text{Nonholonomic constraints: } \theta_{\text{ASV}}(t), \\ \text{Obstacle avoidance for ASV, AUV, and tether,} \\ \text{Tether integrity: } l \geq \|P_{\text{AUV}}(t) - P_{\text{ASV}}(t)\|, \\ \text{Given start/end points for ASV and AUV.} \end{cases}$

In this formulation, $P_{\text{ASV}}(t)$ and $P_{\text{AUV}}(t)$ represent the desired paths over time for the ASV and AUV, respectively.

### D. Trajectory Generation Using Bézier Curves

Bézier curves are employed to generate smooth trajectories between a sequence of waypoints for each robot. The curve for a given path segment is mathematically represented as:

$$B(t) = \sum_{i=0}^{n} \binom{n}{i}(1-t)^{n-i}t^i P_i, \quad (1)$$

where $B(t)$ is the position on the Bézier curve at a parameter $t \in [0, 1]$, $\binom{n}{i}$ denotes the binomial coefficient, $P_i$ are the control points defining the shape of the curve, and $n$ is the degree of the Bézier curve, which is one less than the number of control points. The Bernstein basis polynomials, $b_n^i(t) = \binom{n}{i}(1-t)^{n-i}t^i$, ensure the curve's continuity and smoothness across each segment.

For trajectory planning, this formulation allows the generation of trajectories by interpolating between waypoints using the Bézier curve's control points. This approach provides a way to create smooth and continuous paths for robots by adjusting the control points $P_i$, which are strategically placed to guide the robots' movements from their start positions to their target destinations.

*E. Synchronization of Speeds*

To synchronize the robots, ensuring they reach each waypoint concurrently, the time $t(d)$ to traverse a distance $d$ with acceleration $a$ and maximum speed $v_{max}$ is calculated as follows:

$$t(d) = \begin{cases} \sqrt{\frac{2d}{a}}, & \text{if } d \leq d_{\text{crit}}, \\ \frac{v_{\max}}{a} + \frac{d - d_{\text{accel}}}{v_{\max}}, & \text{otherwise}, \end{cases} \quad (2)$$

where $d_{\text{crit}}$ is the critical distance at which the maximum speed is achieved, defined as $d_{\text{crit}} = \frac{v_{\max}^2}{a}$, and $d_{\text{accel}}$ is the distance travelled during acceleration to $v_{\max}$, given by $d_{\text{accel}} = \frac{1}{2} a \left(\frac{v_{\max}}{a}\right)^2$.

The synchronization time $T_{sync}$ for each waypoint is the maximum of the times calculated for all robots:

$$T_{sync} = \max_{1 \leq i \leq n} T_i, \quad (3)$$

where $T_i$ is the time for the $i$-th robot to reach the waypoint, and $n$ is the total number of robots.

*F. Speed Adjustment for Synchronization*

The speed $V_i$ for the $i$-th segment of each robot's trajectory is adjusted for synchronization:

$$V_i = \frac{d_i}{T_{sync,i} - T_{sync,i-1}}, \quad (4)$$

where $d_i$ is the segment distance, and $T_{sync,i} - T_{sync,i-1}$ is the synchronized time interval for that segment, ensuring temporal coordination between the ASV and AUV for efficient path following.

Our objective is to minimize the objective function $J$, which incorporates terms for velocity $\left|\dot{\mathbf{B}}_i^j(t)\right|^2$ and acceleration $\left|\ddot{\mathbf{B}}_i^j(t)\right|^2$ smoothness, weighted by coefficients $\alpha$ and $\beta$ respectively, subject to non-collision and dynamic constraints:

$$\min J = \alpha \sum_{i=1}^{n_{\text{opt}}} \int_{T_{i-1}}^{T_i} \left|\dot{\mathbf{B}}_i^j(t)\right|^2 dt + \beta \sum_{i=1}^{n_{\text{opt}}} \int_{T_{i-1}}^{T_i} \left|\ddot{\mathbf{B}}_i^j(t)\right|^2 dt,$$

$$\mathbf{B}_i^j \in \text{Obstacle-Free Space},$$
$$-v_{\max} \leq \dot{\mathbf{B}}_i^j(t) \leq v_{\max},$$
$$-a_{\max} \leq \ddot{\mathbf{B}}_i^j(t) \leq a_{\max},$$
$$l \geq \|\mathbf{B}_{\text{AUV}}(t) - \mathbf{B}_{\text{ASV}}(t)\|.$$

With this formulation, our method is able to compensate for the speed variations ensuring synchronization and, thus safe operations of tethered ASV-AUV systems.

**Algorithm 1** CoralGuide algorithm for Tethered ASV-AUV with Nonholonomic Constraints

**Require:** $ASV_{\text{start}}$, $ASV_{\text{target}}$, $AUV_{\text{start}}$, $AUV_{\text{target}}$, ObsMap, TetherLen, MaxSpd, MaxAcc, NhConst, FirstRobot
**Ensure:** $ASV_{\text{Traj}}$, $AUV_{\text{Traj}}$
1: $ASV_{\text{Path}} \leftarrow [\,]$, $AUV_{\text{Path}} \leftarrow [\,]$

**Step 1: Path Generation with A***
2: **if** FirstRobot is ASV **then**
3:     $ASV_{\text{Path}} \leftarrow$ A*($ASV_{\text{start}}$, $ASV_{\text{target}}$, ObsMap, NhConst)
4:     Compute $AUV_{\text{Path}}$ considering $ASV_{\text{Path}}$ and tether
5: **else**
6:     $AUV_{\text{Path}} \leftarrow$ A*($AUV_{\text{start}}$, $AUV_{\text{target}}$, ObsMap)
7:     $ASV_{\text{Path}} \leftarrow$ ComputePath($ASV$, $AUV_{\text{Path}}$, Tether, NhConst)
8: **end if**

**Step 2: Path Smoothing with Bezier Interpolation**
9: $BezierWayPnt \leftarrow [\,]$
10: **for** each segment $S_j$ in combined paths **do**
11:     Calculate control points for $B_j$
12:     Sample points from $B_j$ to generate waypoints
13:     Append waypoints to $BezierWayPnt$
14: **end for**

**Step 3: Trajectory Optimization and Synchronization**
15: $(OptASV_{\text{Traj}}, OptAUV_{\text{Traj}}) \leftarrow$ OptAndSyncTraj(BezierWayPnt, TetherLen, MaxSpd, MaxAcc)

**Step 4: Velocity Smoothing with Cubic Spline**
16: **for** each robot in the system **do**
17:     Apply cubic spline smoothing to velocity profile
18: **end for**

19: **return** $(OptASV_{\text{Traj}}, OptAUV_{\text{Traj}})$

*G. Algorithm*

The full pseudo code of CoralGuide is presented in Algorithm 1. The algorithm's core relies on an obstacle map (*ObsMap*) to guide the selection of viable paths, ensuring obstacle avoidance. Our algorithm is able to initiate the path planning with the ASV or AUV (*FirstRobot*), which gives versatility in operations where one of the vehicles might have priority. The paths are created for both the ASV and AUV while complying with the tether's requirements and adhering to predefined spatial and operational constraints.

Subsequent path refinement employs Bezier interpolation (*BezierWayPnt*), smoothing trajectories to enable practical, controlled navigation. This step is critical for effectively managing the tether's influence on vehicle movement. A significant enhancement in our approach is the trajectory optimization and synchronization phase (*OptAndSyncTraj*). This phase fine-tunes paths to align with maximum speed (*MaxSpd*) and acceleration (*MaxAcc*) constraints while preserving tether integrity. As a result, it achieves synchronized movement between the ASV and AUV, accommodating the tether's length (*TetherLen*) and the dynamic requirements of the system. The final step involves velocity smoothing through cubic spline interpolation for each robot's velocity profile, essential for smoothing speed transitions and enhancing system performance.

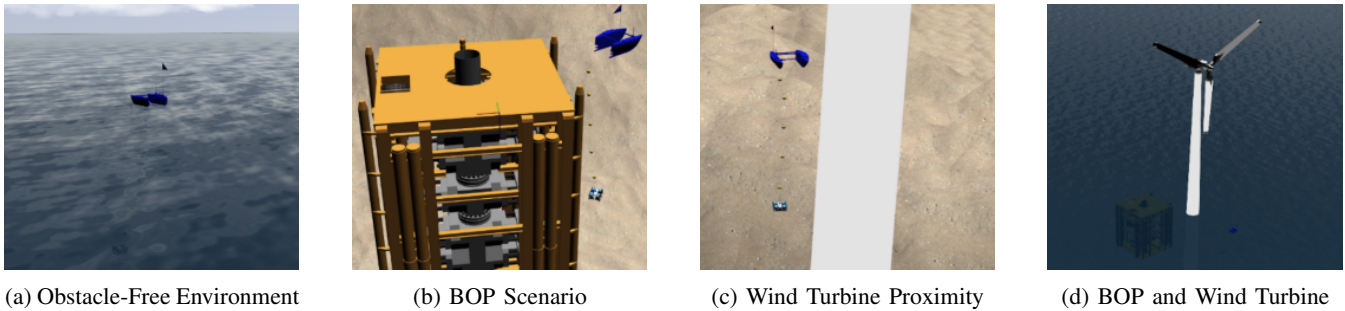

Fig. 3: Overview of test scenarios evaluated.

TABLE I: AVERAGE PERFORMANCE EVALUATION OF ABLATION VARIANTS OF A* ALGORITHM FOR ASV OR AUV INITIATED PATH PLANNING IN OCEAN SCENARIOS

| SCEN | ALGO | LASV | LAUV | NC | CASV | CAUV | AVGD | STDS | STDU | STDT | DOBS | RFRQ |
|---|---|---|---|---|---|---|---|---|---|---|---|---|
| 1 | CoralGuide | 6.053 | 9.146 | – | 2.822 | 11.374 | 5.354 | 0.007 | 0.01 | 0.041 | NA | 9.2 |
| 1 | A* | 6.053 | 9.146 | – | 2.77 | 11.152 | 5.354 | 0.096 | 0.151 | 0.13 | NA | - |
| 2 | CoralGuide | 3.461 | 6.0 | 0 | 0.228 | 0.81 | 5.863 | 0.007 | 0.011 | 0.031 | 0.261 | 18.0 |
| 2 | A* | 3.716 | 6.346 | 8 | 0.195 | 0.67 | 5.927 | 0.067 | 0.128 | 0.104 | 0.103 | - |
| 3 | CoralGuide | 3.79 | 7.737 | 0 | 0.209 | 1.306 | 5.429 | 0.008 | 0.016 | 0.023 | 0.132 | 35.6 |
| 3 | A* | 3.79 | 7.737 | 7 | 0.208 | 1.289 | 5.429 | 0.06 | 0.146 | 0.111 | 0.099 | - |
| 4 | CoralGuide | 6.128 | 8.446 | 0 | 3.262 | 8.854 | 5.293 | 0.009 | 0.014 | 0.042 | 0.343 0.136 | 28.2 |
| 4 | A* | 6.128 | 8.446 | 10 | 3.257 | 8.842 | 5.293 | 0.088 | 0.141 | 0.136 | 0.487 0.169 | - |

## V. EXPERIMENTAL SETUP

For our experiments, we selected two types of vehicles: the BlueBoat [26], and the AUV BlueROV2 [26]. The BlueBoat (ASV) is a nonholonomic robot equipped with two thrusters and GPS to determine its position. The BlueROV2 (AUV) has eight thrusters, and it is equipped with an IMU and a Doppler Velocity Log (DVL) for accurate positioning. For our simulation results, we utilize the UUV simulator [27], [28]. The real-world experiments were conducted in a tank of 12 × 10 × 3 m. For these experiments, we utilised two BlueROV2s. One of the BlueROV2s was modified to move as an ASV, with nonholonomic motion constraints and limited to moving on the surface of the water only. We utilise a BlueROV2 as an ASV for real-world experiments due to limitations with the GPS signal in the testing facilities. The ASV and AUV were connected by a physical cable with a fixed length, mimicking the tether connection. We evaluated four distinct scenarios in the simulation environment:

(a) *Obstacle-Free Environment Scenario* evaluates the basic efficiency of our path-planning algorithm and tether management in ideal conditions.

(b) *Underwater Blowout Preventer (BOP) Navigation Scenario* examines the accuracy and reliability of navigation algorithms in guiding vehicles to specific underwater objectives while navigating around obstacles.

(c) *Wind Turbine Proximity Scenario* tests the system's adaptability in environments where surface obstructions affect underwater path planning, pertinent to operations near underwater installations.

(d) *Combined BOP and Wind Turbine Scenario* provides a comprehensive test of the system's obstacle navigation capabilities in a complex setting that incorporates the challenges of both scenarios (b) and (c) (illustrated in Fig. 3).

The scenarios were also evaluated with either the ASV or AUV as the first vehicle to launch the path planning, looking to demonstrate the versatility of our method. Additionally, we evaluated the Blowout Preventer (BOP) scenario in the tank.

## VI. SIMULATION RESULTS

We evaluated our method on the four cases described previously. For each of these cases, we ran 10 experiments; during these, the start and the goal for both ASV and AUV were randomized. Additionally, we also modify the sequence in which the vehicles commence the path planning process. Table I presents a summary of the obtained results. We utilised the following metrics: Computation Time for Path ASV (CASV), Computation Time for Path AUV (CAUV), Velocity Standard Deviation for ASV (STDS), Velocity Standard Deviation for AUV (STDU), Velocity Standard Deviation for Tether (STDT), Length of Path for AUV from start to goal (LAUV), Length of Path for ASV from start to goal (LASV), Overall Minimum Distance to Obstacle(s) (DOBS), Replanning Frequency (RFRQ), and Number of Collisions (NC).

Despite comparable path lengths and total path lengths for both ASV and AUV from start to goal, along with similar waypoint numbers between both methods, CoralGuide significantly improves velocity synchronization, achieving substantially lower deviation rates. On the other hand, the ablation method's trajectories, although potentially shorter, result in closer proximity to obstacles, which could potentially undermine mission safety. This is critical, as can be

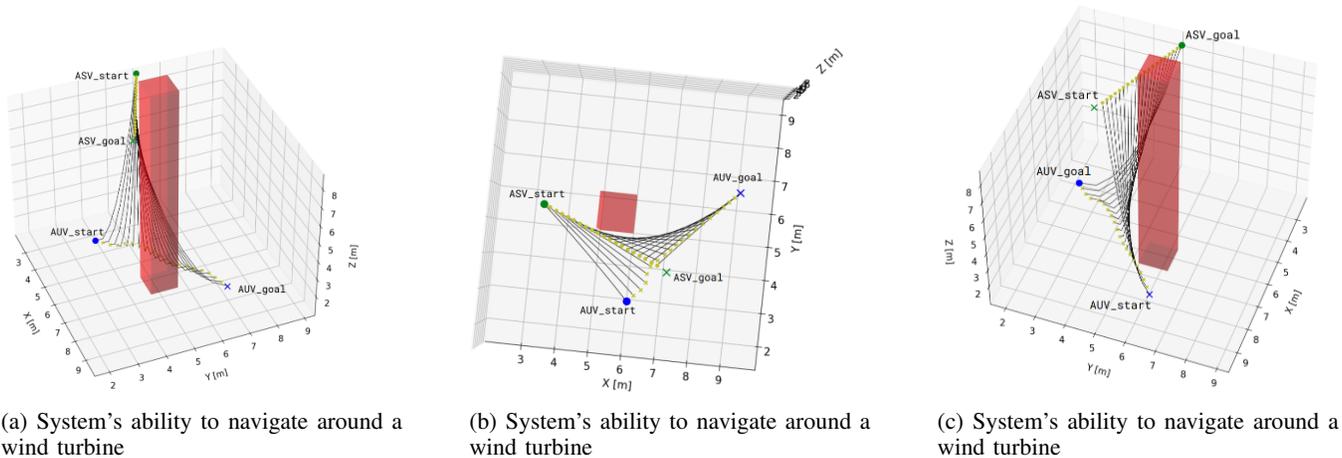

(a) System's ability to navigate around a wind turbine

(b) System's ability to navigate around a wind turbine

(c) System's ability to navigate around a wind turbine

Fig. 4: Overview of CoralGuide's performance for ASV and AUV on Wind Turbine (case c) in simulation.

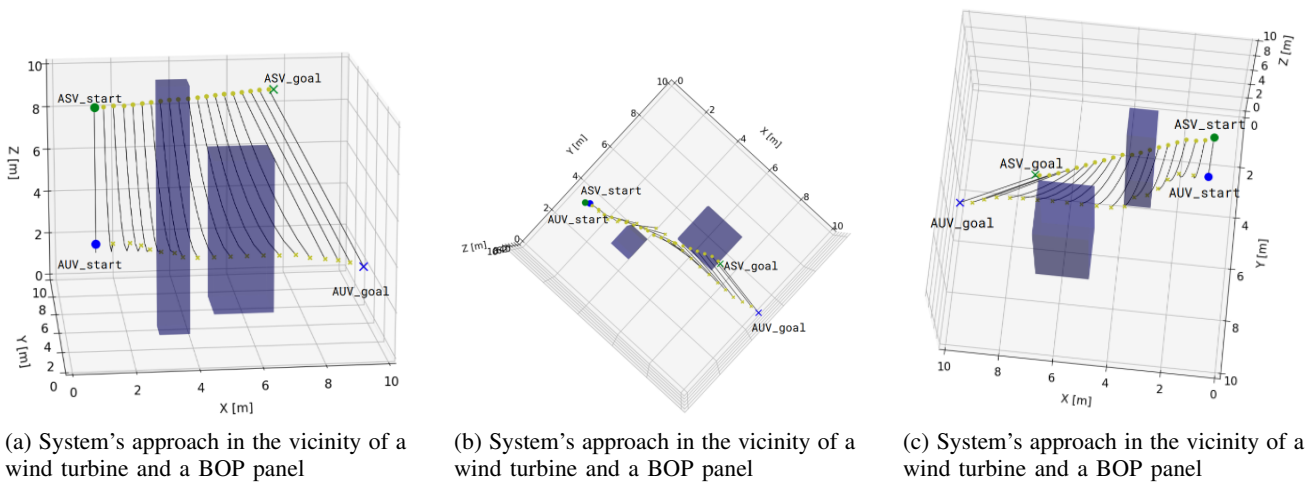

(a) System's approach in the vicinity of a wind turbine and a BOP panel

(b) System's approach in the vicinity of a wind turbine and a BOP panel

(c) System's approach in the vicinity of a wind turbine and a BOP panel

Fig. 5: Overview of CoralGuide's performance for ASV and AUV on BOP and Wind Turbine (case d) in simulation.

seen in Table I, CoralGuide is able to plan and execute extremely complex maneuvers with minimal risk of collisions. This functionality of CoralGuide guarantees that the algorithm consistently identifies collision-free paths from the start to the goal points set by the operator in a multi-agent system environment. This is evidenced by the NC metric, which records zero collisions across all the scenarios we discussed. On the other side, the baseline A* method showed a severe number of collisions, making it unsuitable for this type of scenarios. The baseline A* algorithm demonstrates poor performance, failing to find obstacle-free paths in most cases.

CoralGuide's adaptability is further demonstrated by its capability for frequent replanning to optimize the required tether length. This level of replanning is missing in the ablation approach, highlighting that CoralGuide's comprehensive setup is better equipped to navigate environmental variability and minimize the need for mid-mission adjustments. While the baseline method is quicker in terms of computation times, it compromises essential aspects such as collision prevention and exhibits higher velocity standard deviations. The lack of vehicle synchronization in the baseline A* method leads to instability within the multi-agent system, failing to optimize the tether length, which increases the risk of collisions and path planner failures, making it unsuitable for deployment in even the simplest real-world scenarios.

Furthermore, Figure 4 shows the performance of CoralGuide on the Wind Farm case. It can be seen that our proposed method is able to plan extremely complex manoeuvres while very close to the assets. A more complex case is shown in Figure 5, where the AUV-ASV pair is able to plan around a BOP and Wind Turbine. As can be seen in both figures, these are extremely complex problems that require a high level of synchronization of the multi-robot platforms, both during planning and execution. As such, CoralGuide could enable autonomous multi-robot operations in real-world offshore operations. In the next section, we showcase CoralGuide's performance in a realistic marine operation in a lab environment.

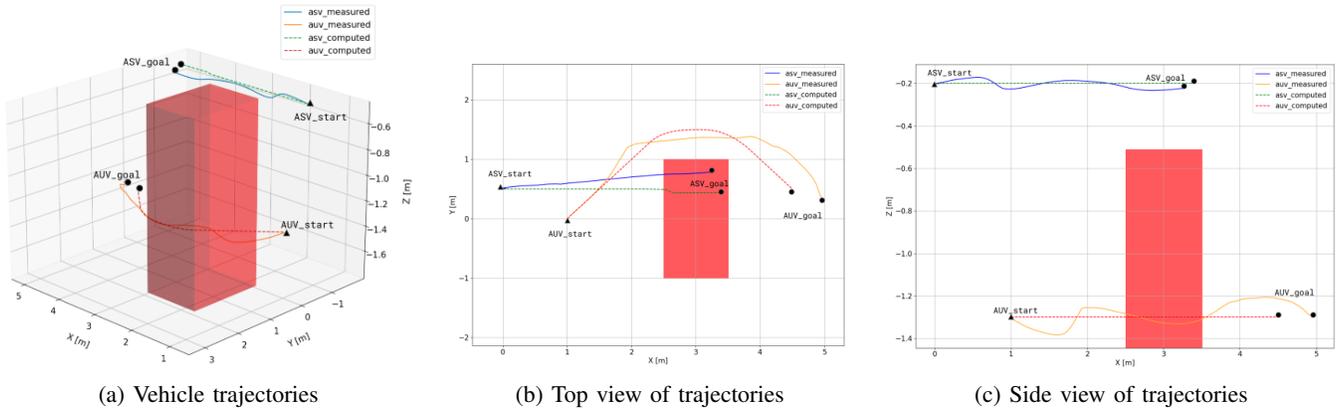

(a) Vehicle trajectories    (b) Top view of trajectories    (c) Side view of trajectories

Fig. 6: CoralGuide's motion performance in the field test scenario.

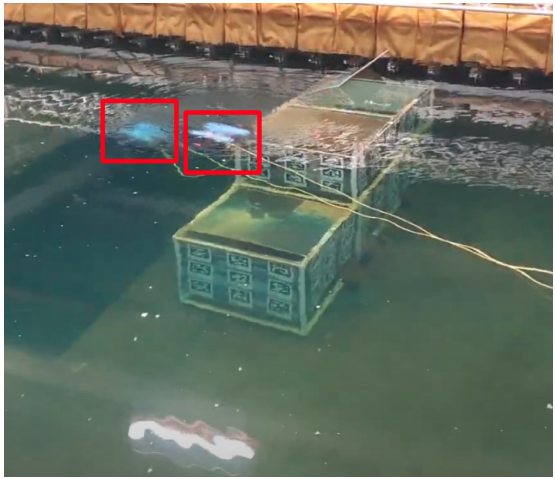

(a) Position of vehicles (indicated by red boxes) in goal position

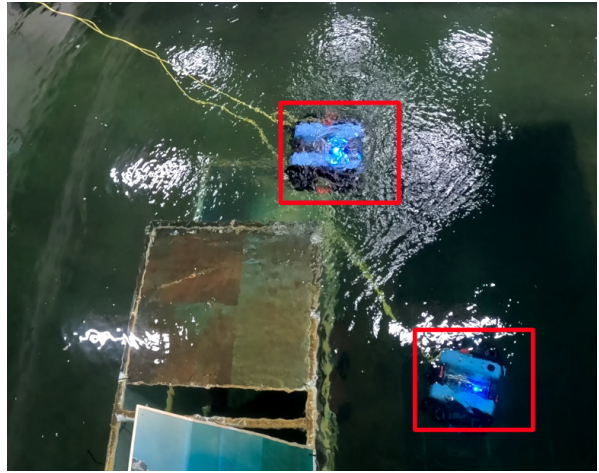

(b) Position of vehicles (indicated by red boxes) in goal position

Fig. 7: Overview of field test scenarios.

TABLE II: PERFORMANCE DURING THE FIELD EXPERIMENTS

| Vehicle | Trajectory Deviation | | |
|---|---|---|---|
| | X [mm] | Y [mm] | Z [mm] |
| ASV | 246 | 345 | 75 |
| AUV | 446 | 827 | 951 |

## VII. FIELD EXPERIMENTS

The conducted field experiments provided insights into the real-world performance of CoralGuide. The obtained results are depicted in Figure 6, while an overview of the testing environment is given in Figure 7. The results confirm the CoralGuide's capability to navigate complex environments, successfully manoeuvre vehicles around obstacles while maintaining a safe distance, and avoid collisions.

The results, detailed in table II, show the average deviations from the planned trajectories, as recorded by the robots' odometry. These deviations are critical indicators of the algorithm's real-world accuracy, highlighting the influence of environmental factors and control system fidelity on the path execution.

Despite the overall success, field experiments showed minor trajectory deviations, with the ASV and AUV experiencing maximum offsets of 246 [mm] and 446 [mm] in the X-axis, 345 [mm] and 857 [mm] in the Y-axis, and 75 [mm] and 951 [mm] in the Z-axis respectively. We believe these variances are due to challenges in the autonomy pipeline, which include the difficulty of accurate pose estimation in marine vehicles, precise low-level control, and dynamic environmental conditions. However, maintaining of a consistent catenary shape without tension, as illustrated in the field tests, highlights the effectiveness of the tether management system. This operational success, coupled with the vehicles' precise goal reach, demonstrates the path-planning algorithm's reliability.

## VIII. CONCLUSION

In this paper we introduced a refined path-planning approach for tethered ASV-AUV systems, enhancing traditional algorithms with dynamic tether management. Our method, CoralGuide, significantly boosts navigation precision, efficiency, and safety. Our method combines the A* algorithm, with catenary tether modelling and Bezier curves, to provide a safe motion planner for tether multi-robot systems. We provided results in simulations, which confirmed the improved performance of vehicle synchronization and tether optimization when utilizing our proposed method. Furthermore, real-world experiments in a water tank validated these results, demonstrating our framework's effectiveness in complex underwater environments, achieving collision-free navigation and upholding path integrity. Future works will look into incorporating advanced control strategies to address multi-body dynamics to further improve our system's adaptability to underwater disturbances and control inputs.

## REFERENCES

[1] Y. Liu, M. Hajj, and Y. Bao, "Review of robot-based damage assessment for offshore wind turbines," *Renewable and Sustainable Energy Reviews*, vol. 158, p. 112187, 2022.
[2] E. Zereik, M. Bibuli, N. Mišković, P. Ridao, and A. Pascoal, "Challenges and future trends in marine robotics," *Annual Reviews in Control*, vol. 46, pp. 350–368, 2018.
[3] G. Battocletti, D. Boskos, D. Tolić, I. Palunko, and B. D. Schutter, "Entanglement definitions for tethered robots: Exploration and analysis," 2024.
[4] V. A. K. T. Rajan, A. Nagendran, A. Dehghani-Sanij, and R. C. Richardson, "Tether monitoring for entanglement detection, disentanglement and localisation of autonomous robots," *Robotica*, vol. 34, no. 3, p. 527–548, 2016.
[5] X. Wang, X. Yao, and L. Zhang, "Path planning under constraints and path following control of autonomous underwater vehicle with dynamical uncertainties and wave disturbances," *Journal of Intelligent & Robotic Systems*, vol. 99, no. 3, pp. 891–908, 2020.
[6] A. Patil, M. Park, and J. Bae, "Coordinating tethered autonomous underwater vehicles towards entanglement-free navigation," *Robotics*, vol. 12, no. 3, 2023.
[7] P. Yao and S. Qi, "Obstacle-avoiding path planning for multiple autonomous underwater vehicles with simultaneous arrival," *Science China Technological Sciences*, vol. 62, no. 1, pp. 121–132, 2019.
[8] S. McCammon and G. A. Hollinger, "Planning and executing optimal non-entangling paths for tethered underwater vehicles," in *2017 IEEE International Conference on Robotics and Automation (ICRA)*, 2017, pp. 3040–3046.
[9] A. Petit, C. Paulo, I. Carlucho, B. Menna, and M. De Paula, "Prediction of the hydrodynamic coefficients of an autonomous underwater vehicle," in *2016 3rd IEEE/OES South American International Symposium on Oceanic Engineering (SAISOE)*, 2016, pp. 1–6.
[10] M. T. Vu, H.-S. Choi, T. Q. M. Nhat, D.-H. Ji, and H.-J. Son, "Study on the dynamic behaviors of an usv with a rov," in *OCEANS 2017 - Anchorage*, 2017, pp. 1–7.
[11] Ángel Madridano, A. Al-Kaff, D. Martín, and A. de la Escalera, "Trajectory planning for multi-robot systems: Methods and applications," *Expert Systems with Applications*, vol. 173, p. 114660, 2021.
[12] L. Antonyshyn, J. Silveira, S. Givigi, and J. Marshall, "Multiple mobile robot task and motion planning: A survey," *ACM Comput. Surv.*, vol. 55, no. 10, feb 2023.
[13] R. Cui, Y. Li, and W. Yan, "Mutual information-based multi-auv path planning for scalar field sampling using multidimensional rrt*," *IEEE Transactions on Systems, Man, and Cybernetics: Systems*, vol. 46, no. 7, p. 993 – 1004, 2016.
[14] J. S. Willners, L. Toohey, and Y. Petillot, "Sampling-based path planning for cooperative autonomous maritime vehicles to reduce uncertainty in range-only localization," *IEEE Robotics and Automation Letters*, vol. 4, no. 4, pp. 3987–3994, 2019.
[15] A. Borgese, D. C. Guastella, G. Sutera, and G. Muscato, "Tether-based localization for cooperative ground and aerial vehicles," *IEEE Robotics and Automation Letters*, vol. 7, no. 3, pp. 8162–8169, 2022.
[16] S. Martínez-Rozas, D. Alejo, F. Caballero, and L. Merino, "Path and trajectory planning of a tethered uav-ugv marsupial robotic system," *IEEE Robotics and Automation Letters*, vol. 8, no. 10, pp. 6475–6482, 2023.
[17] S. Martinez-Rozas, D. Alejo, F. Caballero, and L. Merino, "Optimization-based trajectory planning for tethered aerial robots," in *2021 IEEE International Conference on Robotics and Automation (ICRA)*. IEEE, 2021, pp. 362–368.
[18] M. Cao, K. Cao, S. Yuan, K. Liu, Y. Wong, and L. Xie, "Path planning for multiple tethered robots using topological braids," in *Robotics: Science and Systems XIX*. Robotics: Science and Systems Foundation, Jul. 2023.
[19] M. Cao, K. Cao, S. Yuan, T.-M. Nguyen, and L. Xie, "Neptune: Nonentangling trajectory planning for multiple tethered unmanned vehicles," *IEEE Transactions on Robotics*, vol. 39, no. 4, pp. 2786–2804, 2023.
[20] T. I. Fossen, *Handbook of marine craft hydrodynamics and motion control*. John Wiley & Sons, 2011.
[21] C. Viel, "Self-management of the umbilical of a rov for underwater exploration," *Ocean Engineering*, vol. 248, p. 110695, 2022.
[22] O. Tortorici, C. Péraud, C. Anthierens, and V. Hugel, "Automated deployment of an underwater tether equipped with a compliant buoy-ballast system for remotely operated vehicle intervention," *Journal of Marine Science and Engineering*, vol. 12, no. 2, 2024.
[23] O. Tortorici, C. Anthierens, V. Hugel, and H. Barthelemy, "Towards active self-management of umbilical linking ROV and USV for safer submarine missions," *IFAC-PapersOnLine*, vol. 52, no. 21, pp. 265 – 270, Dec. 2019.
[24] M. Laranjeira, C. Dune, and V. Hugel, "Catenary-based visual servoing for tether shape control between underwater vehicles," *Ocean Engineering*, vol. 200, p. 107018, 2020.
[25] J. Drupt, C. Dune, A. I. Comport, S. Sellier, and V. Hugel, "Inertial-Measurement-Based Catenary Shape Estimation of Underwater Cables for Tethered Robots," in *IEEE/RSJ International Conference on Intelligent Robots and Systems (IROS)*, Kyoto, Japan, Oct. 2022.
[26] Blue Robotics, "Blue robotics – high-quality marine robotics components," https://bluerobotics.com/, 2024.
[27] M. M. M. Manhães, S. A. Scherer, M. Voss, L. R. Douat, and T. Rauschenbach, "UUV simulator: A gazebo-based package for underwater intervention and multi-robot simulation," in *OCEANS 2016 MTS/IEEE Monterey*. IEEE, sep 2016.
[28] Open Source Robotics Foundation, "Gazebo: A multi-robot simulator for outdoor environments," http://gazebosim.org, 2018.